\documentclass[runningheads]{llncs}
\usepackage[T1]{fontenc}
\usepackage{graphicx,verbatim}
\usepackage{booktabs}
\usepackage{float}
\usepackage{makecell}
\usepackage{pifont}
\usepackage{amsmath} 
\usepackage{amssymb}
\usepackage{array}

\usepackage[table]{xcolor}
\usepackage{colortbl}
\definecolor{lightgreen}{HTML}{EBEBEB} 
\definecolor{green}{HTML}{CECECE}      
\definecolor{darkgreen}{HTML}{A4A4A4}  

\usepackage[hidelinks]{hyperref}
\usepackage{cite}

\begin{document}

\title{SG2VID: Scene Graphs Enable Fine-Grained Control for Video Synthesis}

\author{Ssharvien Kumar Sivakumar\inst{1}\orcidID{0009-0009-0383-9462}
Yannik Frisch\inst{1,2} \and
\newline
Ghazal Ghazaei\inst{3} \and
Anirban Mukhopadhyay\inst{1}}

\institute{Technical University Darmstadt, Darmstadt, Germany \email{ssharvien\_kumar.sivakumar@tu-darmstadt.de}
 \and Universitätsmedizin der Johannes Gutenberg-Universität Mainz, Mainz, Germany
 \and Carl Zeiss AG, Munich, Germany}

\maketitle              
\begin{abstract}
Surgical simulation plays a pivotal role in training novice surgeons, accelerating their learning curve and reducing intra-operative errors. However, conventional simulation tools fall short in providing the necessary photorealism and the variability of human anatomy. In response, current methods are shifting towards generative model-based simulators. Yet, these approaches primarily focus on using increasingly complex conditioning for precise synthesis while neglecting the fine-grained human control aspect. To address this gap, we introduce SG2VID, the first diffusion-based video model that leverages Scene Graphs for both precise video synthesis and fine-grained human control. We demonstrate SG2VID's capabilities across three public datasets featuring cataract and cholecystectomy surgery. While SG2VID outperforms previous methods both qualitatively and quantitatively, it also enables precise synthesis, providing accurate control over tool and anatomy's size and movement, entrance of new tools, as well as the overall scene layout. We qualitatively motivate how SG2VID can be used for generative augmentation and present an experiment demonstrating its ability to improve a downstream phase detection task when the training set is extended with our synthetic videos. Finally, to showcase SG2VID's ability to retain human control, we interact with the Scene Graphs to generate new video samples depicting major yet rare intra-operative irregularities. 


\end{abstract}

\section{Introduction}

The Halsted model of "See One, Do One, Teach One" places patients at the centre of novice surgeons' training \cite{wood2022principles}. This often results in suboptimal patient outcomes due to the limited abilities of novice surgeons to handle intra-operative complications \cite{kwong2014long}. Integrating simulation tools into surgical training significantly accelerates their learning curve and reduces the likelihood of procedural mistakes \cite{nair2021effectiveness}. However, existing simulation methods that come in the form of generic phantoms and computer-rendered graphics, lack visual realism and struggle to replicate complex properties of human anatomy \cite{wood2022principles, iliash2024interactive}. To address these issues, recent advancements propose leveraging \textbf{video generative models for surgical simulations} \cite{li2024endora,sun2024bora,zhou2024heartbeat}, which can be tailored around patient-specific conditions, disease severity, and rare or abnormal scenarios.

The generative surgical simulation community predominantly concentrates on adding progressively complex conditioning signals to achieve \textbf{precise synthesis} \cite{iliash2024interactive,zhou2024heartbeat}, often overlooking the crucial element of \textbf{fine-grained human control}. This leads to the perception that achieving precise synthesis inherently comes at the cost of human controllability. We hypothesise that Scene Graphs (SGs) \cite{murali2023latent} can offer a succinct representation to mediate precise synthesis and fine-grained human control. Since SGs efficiently encode surgical scenes by capturing components and their relationships in a structured graphical format, they remain intuitive and human-understandable \cite{holm2023dynamic}. Yet, SGs exhibit strong fine-grained modelling capabilities. For example, SurGrID \cite{frisch2025surgrid} generates new scenes that faithfully adhere to SG conditioning. However, its limitation to generate only static images makes it less suited for surgical simulation, which calls for synthesising temporal sequences.

\begin{figure}
\includegraphics[width=\textwidth]{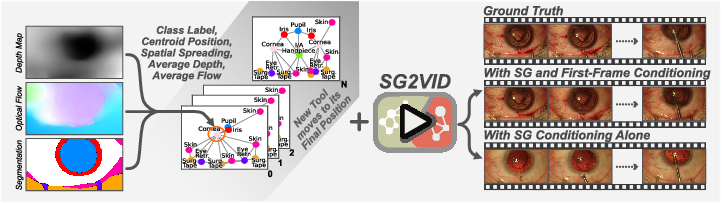}
\caption{\textbf{Surgical Video Synthesis using Scene Graphs (SGs)} and optional first-frame conditioning. Nodes in SGs contain spatial and temporal information about their components, including spatial spreading, centroid position, average depth, and average flow. This enables precise synthesis yet is intuitive for human interaction.} 
\label{fig:concept_figure}
\end{figure}

Endora \cite{li2024endora} and MedSora \cite{wang2024optical} propose unconditional video diffusion for endoscopy simulation. Bora \cite{sun2024bora}, SurGen \cite{cho2024surgen} and VISAGE \cite{yeganeh2024visage} leverage text prompts as conditioning signal from sources like large language models (LLMs) or action graphs, but they fall short in providing precise synthesis. To address this,  Iliash et al. \cite{iliash2024interactive} suggest guiding the generation using a combination of tool masks and text prompts for laparoscopic surgery. Heartbeat \cite{zhou2024heartbeat} incorporates additional structural cues, including sketches, image priors, and optical flow, to further enhance the precision of the synthesis. While complex conditioning signals like masks and optical flow may offer precise control, modifying those to generate new scenes is not a straightforward process. SurgSora \cite{chen2024surgsora} attempts to resolve this by conditioning on user-provided trajectory directions through simple clicks on the first-frame. However, it offers reduced control over the scene composition, such as the size of scene entities, or the entry of new entities, e.g. additional surgical instruments. This is shown in our qualitative comparison in the supplementary. 


We propose \textbf{SG2VID, leveraging SG conditioning for realistic surgical video synthesis with precise and fine-grained human control}. The SGs containing spatial and temporal information, as depicted in Figure \ref{fig:concept_figure}, are used to pre-train two separate graph encoders, capturing local and global features from the surgical scene into an intermediate representation. This representation is subsequently used to condition a video diffusion model \cite{ren2024consisti2v}. The first-frame of any surgical sequence can optionally be utilised as an additional conditioning signal. This is relevant for surgical simulation and analogous to generative editing, where a patient-specific reference image can be used for personalised simulation. In summary, our contributions are as follows: \textbf{(I)} SG2VID is the \textbf{first diffusion-based SG-to-Video model} that enables fine-grained control for video synthesis. The effectiveness of SG2VID is demonstrated across three cataract and cholecystectomy surgery datasets. \textbf{(II)} We showcase SG2VID's generalisation capabilities by generating sequences with precise control from first-frame images of video domains outside of the training data. We also demonstrate how rare samples of intra-operative irregularities can be generated by interactively manipulating the graph, emphasising the human controllability aspect. \textbf{(III)} We highlight our method's potential for generative augmentation by improving a downstream model for surgical phase recognition with the dataset extended with SG2VID's synthetic data.


\section{Method}
We define a Scene Graph as $\mathcal{G} = (\mathcal{V}, \mathcal{E})$, where $\mathcal{V} = \{ v^1, v^2, \dots, v^k \}$ represents a set of $k$ nodes. The nodes are connected by a set of undirected edges $\mathcal{E}$. The Scene Graph $\mathcal{G}$ is constructed using segmentation mask $m$ belonging to an image $x$. Given that the public surgical datasets we utilise provide segmentation annotations for only a small set of videos on limited frames, we employ SASVi \cite{sivakumar2025sasvi} to generate segmentation masks for each frame across all videos. Each node corresponds to a connected component within the mask $m$, such as a tool or anatomical structure. In line with SurGrID \cite{frisch2025surgrid}, each node $v^j \in \mathbb{R}^{d + 7}$ is encoded with a $d$-\textit{dim} class vector, a 2-\textit{dim} centroid position, and 2-\textit{dim} spatial spread representing the height and width of the component. We further extend this representation to include the 2-\textit{dim} average optical flow direction and the 1-\textit{dim} average depth. This additional information assists in modelling the motion trajectory by capturing temporal dynamics within the surgical scene. To compute the average flow and depth values, we leverage pre-trained RAFT \cite{teed2020raft} and MiDaS \cite{ranftl2020towards} models averaged over each component's region. Finally, edges $\mathcal{E}$ in the graph $\mathcal{G}$ connect nodes $\mathcal{V}$ whose corresponding connected components are spatially adjacent in the mask $m$.

\subsection{Pre-Training Graph Encoder}
To enable the graph conditioning of our video diffusion model, we employ two graph encoders, $\mathit{E}_{\mathcal{G}}^{glob}$ and $\mathit{E}_{\mathcal{G}}^{loc}$, which share the same architecture but are trained with different learning objectives. $\mathit{E}_{\mathcal{G}}^{glob}$ captures global scene dynamics, including the overall layout, interactions, and motion patterns, while $\mathit{E}_{\mathcal{G}}^{loc}$ focuses on encoding local information such as the texture and fine-grained details of the components. The graph embeddings from both encoders are concatenated $z_{\mathcal{G}} = concat(z_{\mathcal{G}}^{glob}, z_{\mathcal{G}}^{loc})$ and used to condition the video diffusion model. Each of the graph encoders is constructed using a series of Graph Attention Networks (GATv2) layers \cite{brody2021attentive}, where each node updates its representation by attending to its neighbours using its own representation as the query \cite{brody2021attentive}. A graph-level representation is obtained through mean pooling of the node-level representations.

Given a video sequence consisting of image frames $x_{1:n} = \{x_1, x_2, \dots, x_n\}$ and the corresponding graphs $\mathcal{G}_{1:n} = \{\mathcal{G}_1, \mathcal{G}_2, \dots, \mathcal{G}_n\}$, we randomly select a component, such as a tool or anatomical structure, and mask it across the entire sequence $\tilde{x}_{1:n} = \{\tilde{x}_1, \tilde{x}_2, \dots, \tilde{x}_n\}$. The graph encoder $\mathit{E}_{\mathcal{G}}^{loc}$, together with an auxiliary transformer-based decoder $d_\theta$, is trained to predict the complete sequence $x_{1:n}$ given the masked sequence $\tilde{x}_{1:n}$ and the graph $\mathcal{G}_{1:n}$, by minimizing the reconstruction loss in Equation \ref{eq:local_ge}. This operation takes place in the embedding space, where $z_x = \mathit{E}_x(x)$ and $z_{\tilde{x}} = \mathit{E}_x(\tilde{x})$, but for simplicity, we keep the notations in the equations in the original form.

\begin{equation}
    \label{eq:local_ge}
    \mathcal{L}_{\text{loc}} = \mathbb{E}_{x_{1:n}, \mathcal{G}_{1:n}} \left\| x_{1:n} - d_\theta \left( \tilde{x}_{1:n}, \mathcal{G}_{1:n} \right) \right\|_2^2
\end{equation}
\begin{equation}
    \label{eq:global_ge}
    \mathcal{L}_{\text{glob}} = \mathbb{E}_{\mathcal{G}_{1:n}, m^+_{1:n}, m^-_{1:n}} \left[ -\log \frac{\exp \left(\mathcal{G}_{1:n} \cdot m^+_{1:n} \right)}{\exp \left(\mathcal{G}_{1:n} \cdot m^+_{1:n} \right) + \sum_i \exp \left(\mathcal{G}_{1:n} \cdot m^-_{1:n} \right)} \right]
\end{equation}

In parallel, we train $\mathit{E}_{\mathcal{G}}^{glob}$ using the contrastive learning objective in Equation \ref{eq:global_ge}, aligning the embedding spaces of the graphs $\mathcal{G}_{1:n}$ and segmentation masks $m_{1:n}$. This optimisation is also performed in embedding space, where segmentation masks are encoded using $z_m = \mathit{E}_m(m)$, but the notation is kept in its original form for simplicity. By minimizing Equation \ref{eq:global_ge}, $\mathit{E}_{\mathcal{G}}^{glob}$ is forced to bring compliant target masks $m^+_{1:n}$ closer to $\mathcal{G}_{1:n}$, while pushing the negative target masks $m^-_{1:n}$ further away in the embedding space. By operating in the segmentation space, the model only focuses on capturing the overall layout and ignores textural information, as such details are absent in the masks.

\begin{figure}
\includegraphics[width=\textwidth]{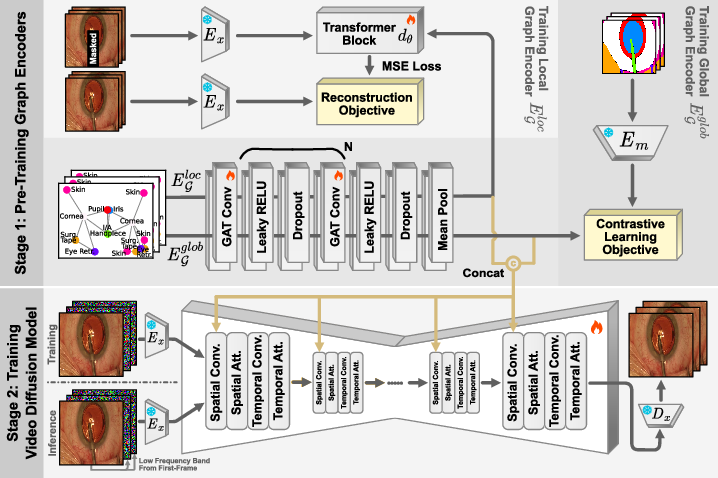}
\caption{Training Workflow of the Graph Encoders and Video Diffusion Model.} 
\label{fig:methodology_figure}
\end{figure}

\subsection{Graph-Image to Video Diffusion Model}
Diffusion Models (DMs) \cite{ho2020denoising} employ a parameterized model $\boldsymbol{\epsilon}_\theta$, typically a 2D U-Net, that learns to model the data distribution by approximating the reversal of gradual noising $p(x_{t-1} | x_t, c)$ by minimizing the error between predicted and target noise: $\min_{\theta}\mathbb{E}_{t, x_0, \epsilon} \left[ \left\| \epsilon - \epsilon_\theta (x_t, t, c) \right\|^2 \right]$. Here, $c$ denotes the conditioning signal, which in this work comprises the graph embeddings $z_{\mathcal{G}_{1:n}}$ and its first-frame $x_1$. For computational efficiency, Rombach et al. \cite{rombach2022high} propose to perform the forward and backward process, $p(z_{t-1} | z_t, c)$, in the latent space. So to apply the first-frame conditioning, we encode the first-frame $z_1 = \mathit{E_x}(x_1)$ and replace it with the first-frame noise $\epsilon^1$. The model input is then constructed as $\hat{\boldsymbol{\epsilon}} = \{ z^1, \epsilon^2, \epsilon^3, \dots, \epsilon^N \}$. To incorporate the graph conditioning signal, the graph embeddings $z_{\mathcal{G}_{1:n}}$ are concatenated with the timestep embedding before being passed to the U-Net blocks. Finally, the sample output from the model can be obtained using a pre-trained decoder $\hat{x} = \mathit{D_x}(\hat{z})$.

For video generation using diffusion models \cite{ho2022video}, the parameterised model is adapted with a spatial-temporal factorised 3D U-Net. A temporal layer, consisting of temporal convolutions and temporal attention, is introduced after each spatial layer. The interleaving of the temporal layers with the existing spatial layers brings temporal awareness to the model. To further improve motion consistency in generated videos and ensure visual coherence from the first-frame throughout the sequence, we incorporate strategies from ConsistI2V \cite{ren2024consisti2v}. Specifically, a local window of first-frame features is incorporated into the temporal layers to augment their attention operations. Common noise schedules have been shown to introduce information leakage about diffusion noise during training, leading to a discrepancy between the initialisation of noise for training and sampling \cite{lin2024common}. To address this issue, ConsistI2V \cite{ren2024consisti2v} integrates the low-frequency components of the first-frame image into the initial noise during inference.

\section{Experiments and Results}
We train and validate our model on three publicly available surgical video datasets featuring real-world cataract and cholecystectomy surgeries: Cataract-1k \cite{ghamsarian2023cataract}, CATARACTS \cite{al2019cataracts}, and Cholec80 \cite{twinanda2016endonet}. These datasets vary significantly in size, containing 1000, 50, and 80 videos, respectively, and also exhibit substantial differences in the average video length. To standardise the training, videos are downsampled to 4 fps, and video sequences of 16 frames are extracted with a 75\% overlap between neighbouring sequences. The frames are downsampled to a resolution of 128×128. The train/validation/test split was assigned on a video-wise basis: 80/10/10\% for Cholec80 and Cataract-1k, and 50/6/44\% for CATARACTS. The graph encoders are trained on a single A40 GPU, while the video diffusion model is trained on four A40 GPUs, though training with a single A40 GPU using gradient accumulation remains feasible. Upon acceptance, we will release our SG representations, model weights, and code at \url{github.com/MECLabTUDA/SG2VID}.

\begin{table}[h!]
    \centering
    \caption{Quantitative Assessment of Video Synthesis Quality.}
    \fontsize{8}{8}\selectfont
    \label{tab:quantitative_result}
    \begin{tabular}{lccc ccc ccc}
        \midrule
        & \multicolumn{3}{c}{\textbf{Cataract-1k} \cite{ghamsarian2023cataract}} & \multicolumn{3}{c}{\textbf{CATARACTS} \cite{al2019cataracts}} & \multicolumn{3}{c}{\textbf{Cholec80} \cite{twinanda2016endonet}} \\
        \cmidrule(lr){2-4} \cmidrule(lr){5-7} \cmidrule(lr){8-10}
        Method & FVD$\downarrow$ & FID$\downarrow$ & LPIPS$\uparrow$ & FVD$\downarrow$ & FID$\downarrow$ & LPIPS$\uparrow$ & FVD$\downarrow$ & FID$\downarrow$ & LPIPS$\uparrow$ \\
        \midrule
        StyleGAN-V \cite{skorokhodov2022stylegan} & 442.6 & 118.2 & 0.286 & \cellcolor{lightgreen}618.7 & 94.8 & 0.382 & 1544.1 & 200.5 & 0.378 \\
        Endora \cite{li2024endora} & \cellcolor{green}265.9 & \cellcolor{green}30.2 & 0.377 & 649.5 & \cellcolor{lightgreen}45.9 & 0.454 & \cellcolor{green}533.8 & \cellcolor{lightgreen}47.0 & 0.525 \\
        MedSora \cite{wang2024optical} & 901.8 & 137.6 & 0.324 & 952.1 & 112.1 & 0.403 & 1297.5 & 153.6 & 0.406 \\
        *LVDM \cite{he2022latent}& 1656.6 & 186.0 & \cellcolor{darkgreen}0.534 & 1178.9 & 113.7 & \cellcolor{darkgreen}0.559 & 1507.0 & 110.8 &  \cellcolor{darkgreen}0.668 \\
        MOFA \cite{niu2024mofa} & 722.2 & 89.9 & 0.361 & 713.4 & 88.3 & \cellcolor{lightgreen}0.460 & 651.1 & 72.4 & 0.506 \\
        \midrule
        SG2VID (Ours) & \cellcolor{darkgreen}77.0 & \cellcolor{darkgreen}15.5 & \cellcolor{lightgreen}0.397 & \cellcolor{darkgreen}523.8 & \cellcolor{green}40.9 & 0.444 & \cellcolor{darkgreen}457.3 & \cellcolor{darkgreen}16.4 &  \cellcolor{lightgreen}0.532 \\
        SG2VID-XIMG (Ours) & \cellcolor{lightgreen}278.4 & \cellcolor{lightgreen}33.3 & \cellcolor{green}0.409 & \cellcolor{green}535.7 & \cellcolor{darkgreen}39.8 & \cellcolor{green}0.465 & \cellcolor{lightgreen}560.1 & \cellcolor{green}25.1 & \cellcolor{green}0.533 \\
        \midrule
        
    \end{tabular}
    \begin{tabular}{lcc cc cc}
        \midrule
        & \multicolumn{2}{c}{\textbf{Cataract-1k} \cite{ghamsarian2023cataract}} & \multicolumn{2}{c}{\textbf{CATARACTS} \cite{al2019cataracts}} & \multicolumn{2}{c}{\textbf{Cholec80} \cite{twinanda2016endonet}} \\
        \cmidrule(lr){2-3} \cmidrule(lr){4-5} \cmidrule(lr){6-7}
        Method & BB IoU$\uparrow$ & F1$\uparrow$ & BB IoU$\uparrow$ & F1$\uparrow$ & BB IoU$\uparrow$ & F1$\uparrow$  \\
        \midrule
        *LVDM \cite{he2022latent} & 0.193 & 0.164 & 0.225 & 0.149 & 0.42 & 0.28 \\
        MOFA \cite{niu2024mofa} & \cellcolor{green}0.456 & \cellcolor{lightgreen}0.449 & \cellcolor{green}0.425 & \cellcolor{lightgreen}0.275 & \cellcolor{green}0.599 & \cellcolor{green}0.45 \\
        \midrule
        SG2VID (Ours) & \cellcolor{darkgreen}0.624 & \cellcolor{darkgreen}0.634 & \cellcolor{darkgreen}0.493 & \cellcolor{darkgreen}0.379 & \cellcolor{darkgreen}0.623 & \cellcolor{darkgreen}0.476 \\
        SG2VID-XIMG (Ours) & \cellcolor{lightgreen}0.440 & \cellcolor{green}0.460 & \cellcolor{lightgreen}0.389 & \cellcolor{green}0.318 & \cellcolor{lightgreen}0.530 & \cellcolor{lightgreen}0.389 \\
        \midrule
        Mask R-CNN on Annotated Subset & 0.731 & 0.745 & 0.636 & 0.585 & 0.887 & 0.957 \\

        \midrule
    \end{tabular}
\end{table}

\textbf{Quantitative Assessment:} To quantitatively assess the quality and diversity of the generated sequences, we utilise a combination of FID, FVD \cite{unterthiner2018fvd} and LPIPS diversity score \cite{zhang2018lpips}, with results shown in Table \ref{tab:quantitative_result} \textit{(top)}. Additionally, to assess how well the generated sequences align with the SG conditioning, we use Mask R-CNN \cite{he2017mask} for component detection on both the generated and real sequences. We adopt this strategy because the bounding box (BB) annotations are limited to a small subset of frames. We therefore train the Mask R-CNN model on this annotated subset. We then compare the model's predictions between the real and generated sequences, evaluating component classification using F1 score and localisation accuracy using BB IoU. The results of this evaluation are reported in Table \ref{tab:quantitative_result} \textit{(bottom)}.

We compare our method against various approaches with publicly available repositories. StyleGAN-V \cite{skorokhodov2022stylegan}, Endora \cite{li2024endora}, and MedSora \cite{wang2024optical} are unconditional video generation models. We adapt LVDM \cite{he2022latent} for text-based conditioning and generate textual descriptions using SG triplets, as in VISAGE \cite{yeganeh2024visage}. For trajectory-based conditioning, MOFA \cite{niu2024mofa} employs a combination of first-frame and motion trajectory, obtained through applying sparse motion sampling on optical flow, similar to the strategy of SurgSora \cite{chen2024surgsora}. As shown in Table \ref{tab:quantitative_result}, SG2VID surpasses all the baselines in terms of visual fidelity, while adhering to the conditioning signal. SG2VID-XIMG is our model trained using graphs alone without the first-frame. 

\textbf{Ablation:} We perform an ablation study on Cataract-1k to highlight the importance of jointly modelling local and global cues. SG2VID's local-only variant scores an FVD of 80.6, FID of 16.4, and LPIPS of 0.396, while the global-only variant scores 79.8, 15.9, and 0.397, both underperforming SG2VID.


\begin{figure}
\centering
\includegraphics[width=\textwidth]{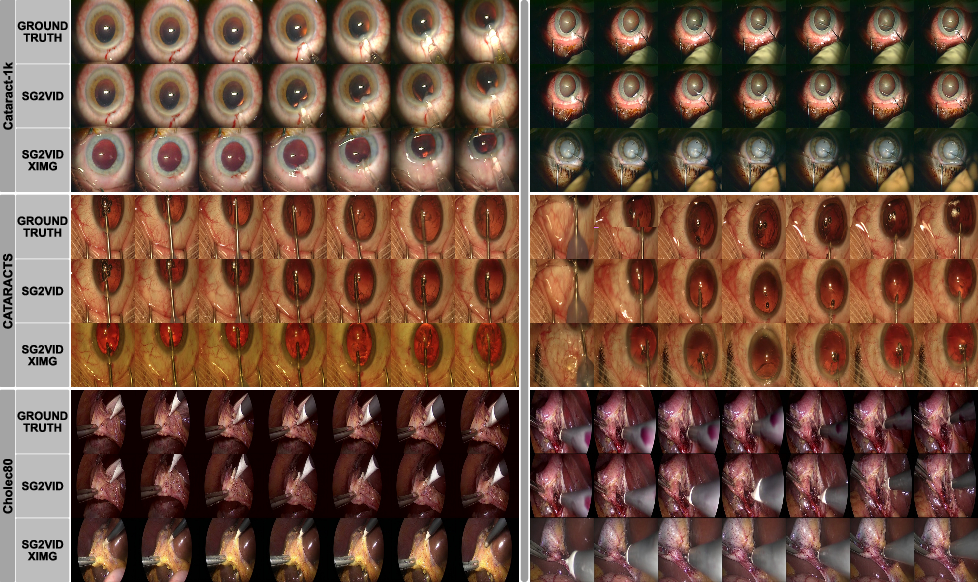}
\caption{\textbf{SG-to-Video Generation}. More samples are in the supplementary material.} 
\label{fig:qualitative_result}
\end{figure}

\textbf{Qualitative Assessment:} Figure \ref{fig:qualitative_result} presents qualitative results of SG2VID and SG2VID-XIMG, demonstrating how SG conditioning enables precise synthesis. In the synthesised sequences, the movement and size of the tools and anatomies, as well as the overall layout, closely follow the ground truth. SG2VID also has no difficulty synthesising the entry of a brand-new tool absent in the first-frame, as well as the exit and reentry of existing tools. Without the first-frame conditioning, SG2VID-XIMG still accurately captures tool and anatomy movements, but video-specific features absent from the SG, such as eye colour, appear random, akin to generative augmentation. We provide additional samples of our synthetic videos in the supplementary material, including qualitative comparisons to related work.

\begin{figure}
\includegraphics[width=\textwidth]{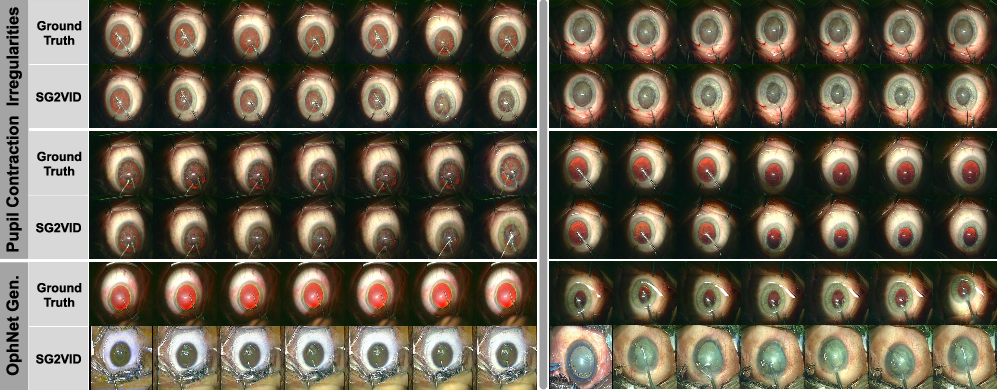}
\caption{Synthesisation of Pupil Contraction Irregularities and Generalisation on Unseen Images from the OphNet Dataset. More samples are in the supplementary material.} 
\label{fig:ophnet_pupilcollapse}
\end{figure}

\textbf{Method Controllability and Generalisation:} To highlight human controllability and generalisation to uncommon cases, we simulate sudden pupil contraction known as miosis, a rare intra-operative irregularity described in Cataract-1k \cite{ghamsarian2023cataract}. This phenomenon risks significant tissue injury, especially when a tool is deeply inserted. We select a set of SGs and reduce the pupil size in the final SG. The pupil size values for the intermediate SGs are simply interpolated between the initial and final SG. Figure \ref{fig:ophnet_pupilcollapse} \textit{(top)} shows synthesised sequences with sudden pupil contractions. To demonstrate generalisation, we generate sequences from entirely unseen eye images to recreate unseen domain style transfer. We take eye images from the OphNet dataset \cite{hu2024ophnet} and condition on SGs from Cataract-1k to synthesise new video sequences, shown in Figure \ref{fig:ophnet_pupilcollapse} \textit{(bottom)}. We also provide more samples for both cases in the supplementary material.

\textbf{Generative Augmentation on Surgical Phase Recognition:} The qualitative results in Figure \ref{fig:qualitative_result} motivate the use of SG2VID as a generative augmenter. We select surgical phase classification on CATARACTS \cite{al2019cataracts} as a downstream task and train on our generatively extended dataset. To extend the training set, we use first-frame from one video and SGs from another (mix-and-match) within the training set of CATARACTS \cite{al2019cataracts} to generate new video sequences. Since our model allows for precise control over most components during the synthesis process, this would be comparable to sequence-to-sequence translation. Given that our generated sequences are only 16 frames long, we generate each subsequent sequence using the last frame of the previous one, producing complete surgical videos from start to finish. Phase annotations for synthesised videos are the same as the target video from which we take the graph. We train MS-TCN++ \cite{li2020ms} model using features from DINO \cite{caron2021emerging} on both the original and the extended datasets containing our generated videos. The macro F1-Score and Accuracy when trained on the normal training set are 0.794 and 0.793, respectively. Extending the training set with our synthesised videos improves the downstream phase recognition, yielding macro F1 and Accuracy scores of 0.805 and 0.816.
\section{Conclusion}
We present the \textbf{first diffusion-based SG-to-Video model (SG2VID)}, enabling fine-grained video synthesis through human-interactable conditioning. Our method works across multiple surgical domains, demonstrating both qualitative and quantitative superiority. SG2VID provides precise control over the movement and size of tools and anatomical structures, as well as new tool entry and exit. To showcase its human controllability, we generate rare pupil collapse irregularities in cataract surgery by interactively manipulating the SGs. Our results highlight SG2VID’s potential as a generative augmenter, improving downstream phase recognition with a generatively extended training set.


\begin{credits}
\subsubsection{\ackname} The work is partially funded by the German Federal Ministry of Education and Research as part of the Software Campus (research grant 01|S23067).

\end{credits}

%
%
%
\bibliographystyle{splncs04}
\bibliography{mybibliography}





\end{document}